\documentclass[lettersize,journal]{IEEEtran}
\usepackage{times}
\usepackage{epsfig}
\usepackage{cite}
\usepackage{graphicx}
\usepackage{amsmath}
\usepackage{amssymb}
\usepackage{array}
\usepackage[export]{adjustbox}

\usepackage[accsupp]{axessibility}  
\usepackage{caption}
\usepackage{subcaption}

\begin{document}
%
\title{The Impact of Partial Occlusion on Pedestrian Detectability}
%
%
%

\author{Shane Gilroy,
        Darragh Mullins,
        Edward Jones, 
        Ashkan Parsi,
        Martin Glavin,
\thanks{S.Gilroy is with the Department of Computing and Electronic Engineering at Atlantic Technological University, Sligo and the Connaught Automotive Research Group, University of Galway, Ireland e-mail: (shane.gilroy@atu.ie).}
\thanks{E.Jones, D.Mullins, A.Parsi and M.Glavin are with the Connaught Automotive Research Group, University of Galway, Ireland.}
}

%
%

\markboth{Published version of this research available at https://doi.org/10.1016/j.birob.2023.100115}
{Shell \MakeLowercase{\textit{et al.}}: Bare Demo of IEEEtran.cls for IEEE Journals}


\maketitle

\begin{abstract}
Robust detection of vulnerable road users is a safety critical requirement for the deployment of autonomous vehicles in heterogeneous traffic. One of the most complex outstanding challenges is that of partial occlusion where a target object is only partially available to the sensor due to obstruction by another foreground object. A number of leading pedestrian detection benchmarks provide annotation for partial occlusion, however each benchmark varies greatly in their definition of the occurrence and severity of occlusion. Recent research demonstrates that a high degree of subjectivity is used to classify occlusion level in these cases and occlusion is typically categorized into 2-3 broad categories such as “partially” and “heavily” occluded. In addition, many pedestrian instances are impacted by multiple inhibiting factors which contribute to non-detection such as object scale, distance from camera, lighting variations and adverse weather. This can lead to inaccurate or inconsistent reporting of detection performance for partially occluded pedestrians depending on which benchmark is used.
This research introduces a novel, objective benchmark for partially occluded pedestrian detection to facilitate the objective characterization of pedestrian detection models. Characterization is carried out on seven popular pedestrian detection models for a range of occlusion levels from 0-99\%, in order to demonstrate the efficacy and increased analysis capabilities of the proposed characterization method. 
Results demonstrate that pedestrian detection performance degrades, and the number of false negative detections increase as pedestrian occlusion level increases. Of the seven popular pedestrian detection routines characterized, CenterNet has the greatest overall performance, followed by SSDlite. RetinaNet has the lowest overall detection performance across the range of occlusion levels. 
\end{abstract}


%
\IEEEpeerreviewmaketitle

\section{Introduction}

\IEEEPARstart{A}{ccurate} and robust pedestrian detection systems are an essential requirement for the safe navigation of autonomous vehicles in heterogeneous traffic.
The SAE J3016 standard \cite{sae2021taxonomy} defines levels of driving automation ranging from Level 0, where the vehicle contains zero automation and the human driver is in complete control, to level 5 where the vehicle is solely responsible for all perception and driving tasks in all scenarios. The progression from automation levels 3-5 requires a significant increase in assumption of responsibility by the vehicle, placing progressively increasing demands on the performance of pedestrian detection systems to inform efficient path planning and to ensure the safety of vulnerable road users.
Despite recent improvements in pedestrian detection systems, many challenges still exist before we reach the object detection capabilities required for safe autonomous driving.
One of the most complex and persistent challenges is that of partial occlusion, where a target object is only partially available to the sensor due to obstruction by another foreground object. The frequency and variety of occlusion types in the automotive environment is large and diverse as pedestrians navigate between vehicles, buildings, traffic infrastructure and other road users. Pedestrians can be occluded by static or dynamic objects, may inter-occlude (occlude one another) such as in crowds, and self-occlude - where parts of a pedestrian overlap.

\begin{table}[]
\caption{Categories of occlusion levels by dataset.}
\resizebox{\linewidth}{!}{%
\begin{tabular}{|l|c|c|c|}
\hline
\textbf{Dataset} & \multicolumn{3}{c|}{\textbf{Occlusion Level}} \\ \cline{2-4} 
 & \textit{Low} & \textit{Partial} & \textit{Heavy} \\ \hline
\begin{tabular}[c]{@{}l@{}}EuroCity\\ Persons \cite{braun2019eurocity}\end{tabular} & \textless{}40\% & 40-80\% & \textgreater{}80\% \\ \hline
CityPersons \cite{zhang2017citypersons} & - & \textless{}35\% & 35-75\% \\ \hline
KITTI \cite{geiger2012we} & \begin{tabular}[c]{@{}c@{}}"Fully \\ Visible"\end{tabular} & \begin{tabular}[c]{@{}c@{}}"Partially \\ Occluded"\end{tabular} & \begin{tabular}[c]{@{}c@{}}"Difficult \\ to See"\end{tabular} \\ \hline
\begin{tabular}[c]{@{}l@{}}Caltech \\ Pedestrian \cite{dollar2009pedestrian}\end{tabular} & - & 1-35\% & 35-80\% \\ \hline
\begin{tabular}[c]{@{}l@{}}Multispectral \\ Pedestrian \cite{hwang2015multispectral}, \\ OVIS \cite{qi2021occluded}\end{tabular} & - & $\leq${}50\% & \textgreater{}50\% \\ \hline
TJU-DHD \cite{pang2020tju} & - & $\leq${}35\% & \textgreater{}35\% \\ \hline
\begin{tabular}[c]{@{}l@{}}Daimler \\ Tsinghua \cite{li2016new}\end{tabular} & \textless{}10\% & 10-40\% & 41-80\% \\ \hline

\hline

\end{tabular}%
}
\end{table}

Leading pedestrian detection systems claim a detection performance of approximately 65\%-75\% of partially and heavily occluded pedestrians respectively using current benchmarks \cite{gilroy2019overcoming}\cite{ning2021survey}\cite{cao2021handcrafted}\cite{xiao2021deep}. However, recent research \cite{gilroy2021pedestrian} demonstrates that the definition of the occurrence and severity of occlusion varies greatly, and a high degree of subjectivity is used to categorize pedestrian occlusion level in each benchmark. Occlusion is typically split into 2-3 broad, loosely defined, categories such as “partially” or “heavily” occluded, Table \ref{tab:datasets} \cite{gilroy2021pedestrian}. In addition, many pedestrian instances are impacted by multiple inhibiting factors which contribute to non-detection such as object scale, distance from camera, lighting variations and adverse weather. This makes it difficult to determine if the primary factor for non-detection is the severity of occlusion alone and can lead to inaccurate or inconsistent reporting of detection performance for partially occluded pedestrians depending on which benchmark is used.
A knowledge gap exists for objective, detailed occlusion level analysis for pedestrian detection across the complete spectrum of occlusion levels. Use of an objective, fine grained occlusion specific benchmark will result in more objective, consistent and detailed analysis of pedestrian detection algorithms for partially occluded pedestrians.

This research proposes a novel, objective benchmark for partially occluded pedestrian detection to facilitate the objective characterization of pedestrian detection models. Objective characterization of occluded pedestrian detection performance is carried out for seven popular pedestrian detection routines for a range of occlusion levels from 0-99\%.
The contributions of this research are:
1. A novel, objective, test benchmark for partially occluded pedestrian detection is presented.
2. Objective characterization of pedestrian detection performance is carried out for seven popular pedestrian detection routines.



\begin{table*}[]
\caption{Overview of Pedestrian Detection Models.}
\label{table:models}
\begin{center}
\begin{tabular}{|l|c|c|c|c|}
\hline
\textbf{Model} & \textbf{Classifier} & \textbf{Training Data} & \textbf{Weights Source} & \textbf{Performance (mAP)} \\ \hline\hline
\textbf{FasterRCNN} \cite{ren2016faster} & ResNet-50 FPN & COCO & Voxel51\cite{voxel51_mz} & 0.398 \\ \hline
\textbf{MaskRCNN} \cite{he2017mask} & ResNet-50 FPN & COCO & Voxel51\cite{voxel51_mz} & 0.411 \\ \hline
\textbf{R-FCN} \cite{dai2016r} & ResNet-101 & COCO & Voxel51\cite{voxel51_mz} & 0.411 \\ \hline
\textbf{SSD} \cite{liu2016ssd} & VGG16 & COCO & Torchvision\cite{torchvision_mz} & 0.412 \\ \hline
\textbf{SSDlite} \cite{howard2019searching} \cite{sandler2018mobilenetv2} & MobileNetV3 Large & COCO & Torchvision\cite{torchvision_mz} & 0.464 \\ \hline
\textbf{RetinaNet} \cite{lin2017focal} & ResNet-50 FPN & COCO & Voxel51\cite{voxel51_mz} & 0.361 \\ \hline
\textbf{CenterNet} \cite{zhou2019objects} & Hourglass-104 & COCO & Voxel51\cite{voxel51_mz} & 0.533 \\ \hline

\hline
\end{tabular}
\end{center}
\end{table*}


\begin{figure*}[t]
\begin{center}
   \includegraphics[width=0.85\textwidth]{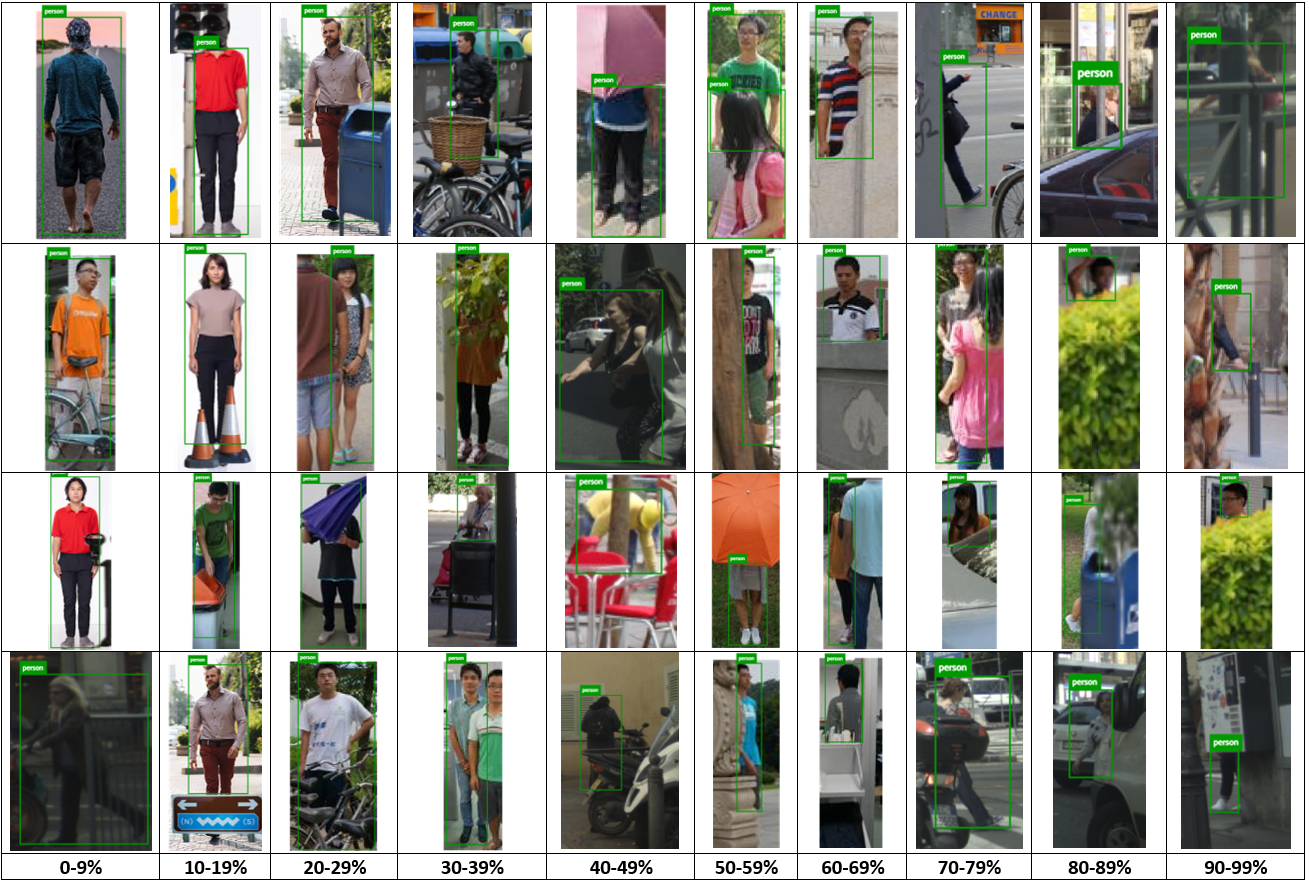}
\end{center}
   \caption{Dataset Sample. An example of dataset images for each level of occlusion. The custom dataset contains 820 pedestrian instances containing a wide range of pedestrian poses and occluding objects. All images are compiled from publicly available sources.}
\label{fig:ds_sample}
\end{figure*}


\begin{figure}[]
\begin{center}
   \includegraphics[width=\linewidth]{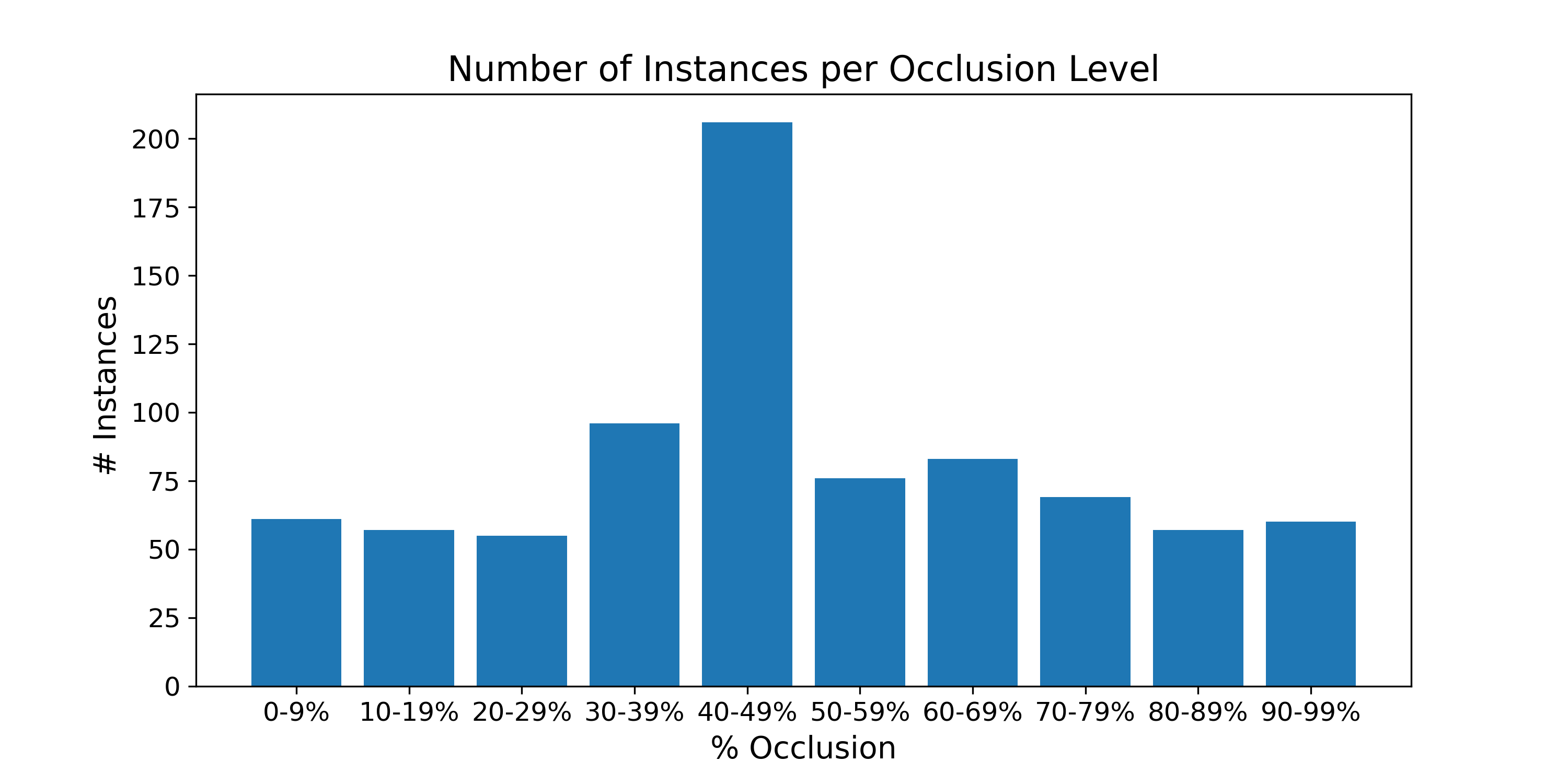}
\end{center}
   \caption{Dataset Statistics. The number of pedestrian instances per occlusion level. The custom dataset contains 820 pedestrian instances under progressive levels of occlusion from 0-99\%.}
\label{fig:ds_stats}
\end{figure}









\section{Related Work}

A number of popular pedestrian detection benchmarks provide annotation of pedestrian occlusion level to determine the relative detection performance for partially occluded pedestrians.
Dollar et al \cite{dollar2011pedestrian} provides analysis on occluded pedestrians based on the Caltech Pedestrian Dataset \cite{dollar2009pedestrian}. Caltech Pedestrian estimates the occlusion ratio of pedestrians by annotating 2 bounding boxes, one for the visible pedestrian area and one for the annotators estimate of the total pedestrian area. Pedestrians are categorised into 2 occlusion categories, “partially occluded”, defined as 1-35\% occluded and “heavily occluded”, defined as 35-80\% occluded. Any pedestrians suspected to be more that 80\% occluded are labelled as fully occluded. Analysis of the frequency of occlusion on the Caltech Pedestrian Dataset demonstrated that over 70\% of pedestrians were occluded in at least one frame, highlighting the frequency of occurrence of pedestrian occlusion in the automotive environment. 
The Eurocity Persons \cite{braun2019eurocity} Dataset categorizes pedestrians according to three occlusion levels: low occlusion (10\%-40\%), moderate occlusion (40\%-80\%), and strong occlusion (larger than 80\%). Classification is carried out by human annotators in a similar manner to the Caltech Pedestrian Dataset. The full extent of the occluded pedestrian is estimated, and the approximate level of occlusion is then estimated to be within one of the three defined categories.
Citypersons \cite{zhang2017citypersons} calculate occlusion levels by drawing a line from the top of the head to the middle of the two feet of the occluded pedestrian. Human annotators are required to estimate the location of the head and feet if these are not visible. A bounding box is then generated for the estimated full pedestrian area using a fixed aspect ratio of 0.41(width/height). This is then compared to the visible area bounding box to denote occlusion level. These estimates of occlusion level are then categorized into two levels, “reasonable” ($<$=35\% occluded) and “heavy occlusion” (35\%-75\%). Similar approaches are taken in \cite{pang2020tju}\cite{zhang2016far}\cite{shao2018crowdhuman}\cite{chi2020pedhunter} \cite{li2016unified}.
The Kitti Vision Benchmark \cite{geiger2012we} and Multispectral Pedestrian Dataset \cite{hwang2015multispectral} tasked human annotators with marking each pedestrian bounding box as “visible”, “semi-occluded”, “fully-occluded”.
Although these methods are useful for the relative comparison of detection performance on specific datasets, the occlusion categories used are broad (usually 2 to 3 categories), are inconsistent from benchmark to benchmark, and involve a high degree of subjectivity by the human annotator, Table \ref{tab:datasets} \cite{gilroy2021pedestrian}\cite{gilroy2022objective}. A knowledge gap exists for a detailed, objective benchmark to compare pedestrian detection performance for partially occluded pedestrians.
Many pedestrian detection analysis papers \cite{dollar2011pedestrian}\cite{zhang2016far}\cite{walk2010new}\cite{rajaram2015exploration}\cite{mao2017can}\cite{zhang2017towards}\cite{ragesh2019pedestrian}\cite{cao2019taking}\cite{toprak2020limitations}\cite{hasan2021generalizable} and occlusion-specific survey papers \cite{gilroy2019overcoming}\cite{ning2021survey}\cite{xiao2021deep}\cite{chandel2015occlusion}\cite{saleh2021occlusion} highlight the outstanding challenges posed by occluded pedestrians, however, no known objective characterization of pedestrian detection performance spanning the spectrum of occlusion levels has been carried out to date.

Gilroy et al \cite{gilroy2022objective} describes an objective method of occlusion level annotation and visible body surface area estimation of partially occluded pedestrians. Keypoint detection is applied to identify semantic body parts and findings are cross-referenced with a visibility score and the pedestrian mask in order to confirm the presence or occlusion of each semantic part. A novel method of 2D body surface area estimation based on the "Wallace rule of Nines" \cite{gilroy2021pedestrian}\cite{wallace1951exposure} is then used to quantify the total occlusion level of pedestrians. 
Experimental results demonstrate that the proposed objective occlusion level classifier outperforms prior works and more closely matches the pixel wise occlusion level of pedestrians in images than the previous state of the art.


\begin{figure}[h!]
\begin{center}
    \includegraphics[width=0.305\linewidth]{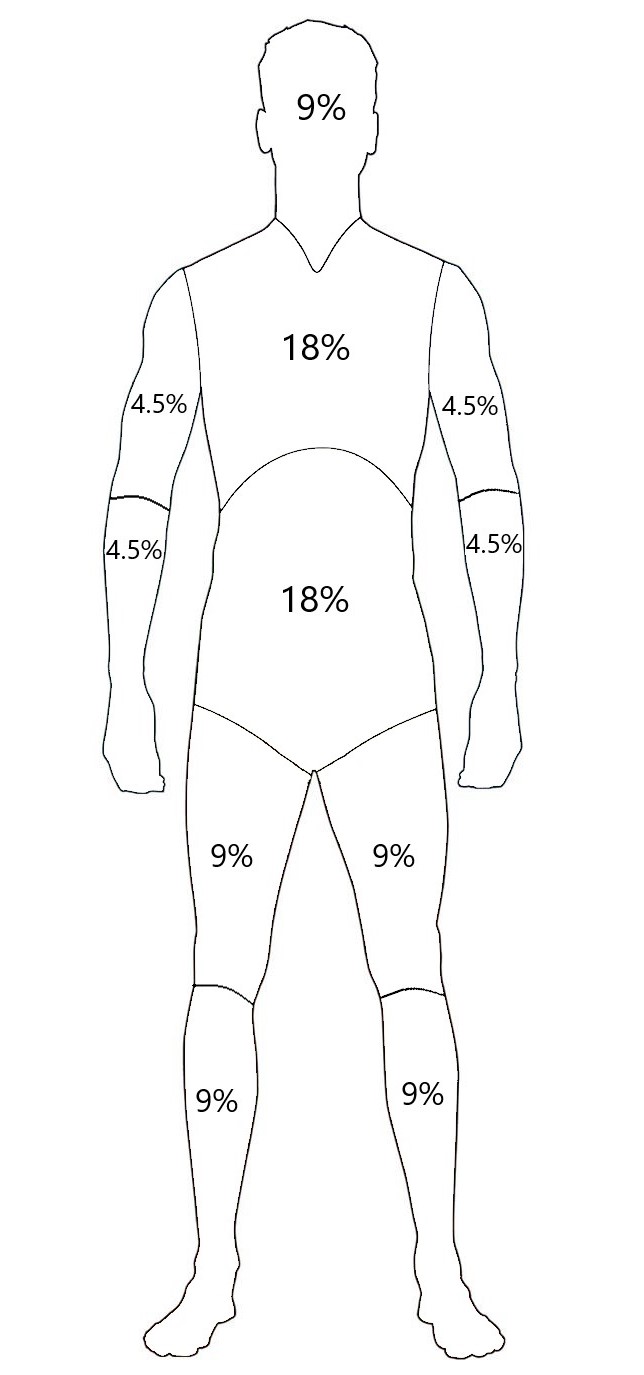}
\end{center}
   \caption{2D Body Surface Area \cite{gilroy2022objective}}
\label{fig:2DBSA}
\end{figure}


\begin{figure*}[]
\begin{center}
    \includegraphics[width=\textwidth]{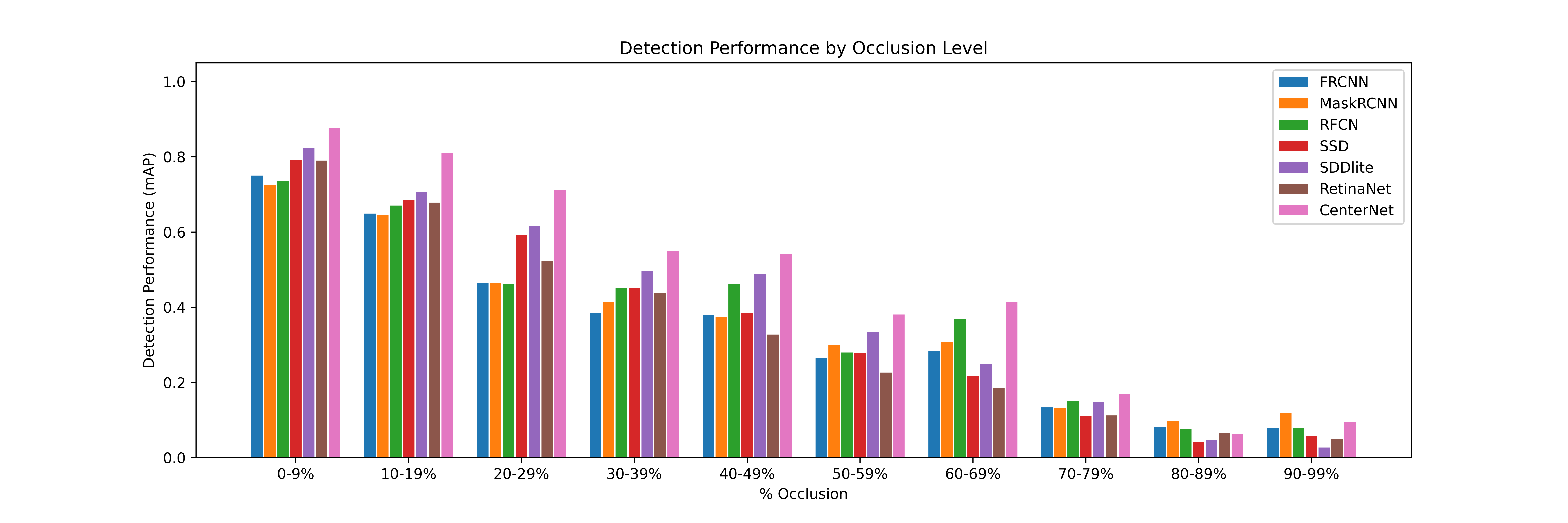}
\end{center}

\caption{Detection Performance by Occlusion Level. Pedestrian detection performance of seven popular pedestrian detection models is displayed for images containing progressive levels of occlusion. Pedestrian detection performance declines as the level of pedestrian occlusion is increased. CenterNet\cite{zhou2019objects} is the highest performing detection model for pedestrians up to 80\% occluded.}
\label{fig:OccVsDet}
\end{figure*}


\begin{figure}[]
\begin{center}
   \includegraphics[width=\linewidth]{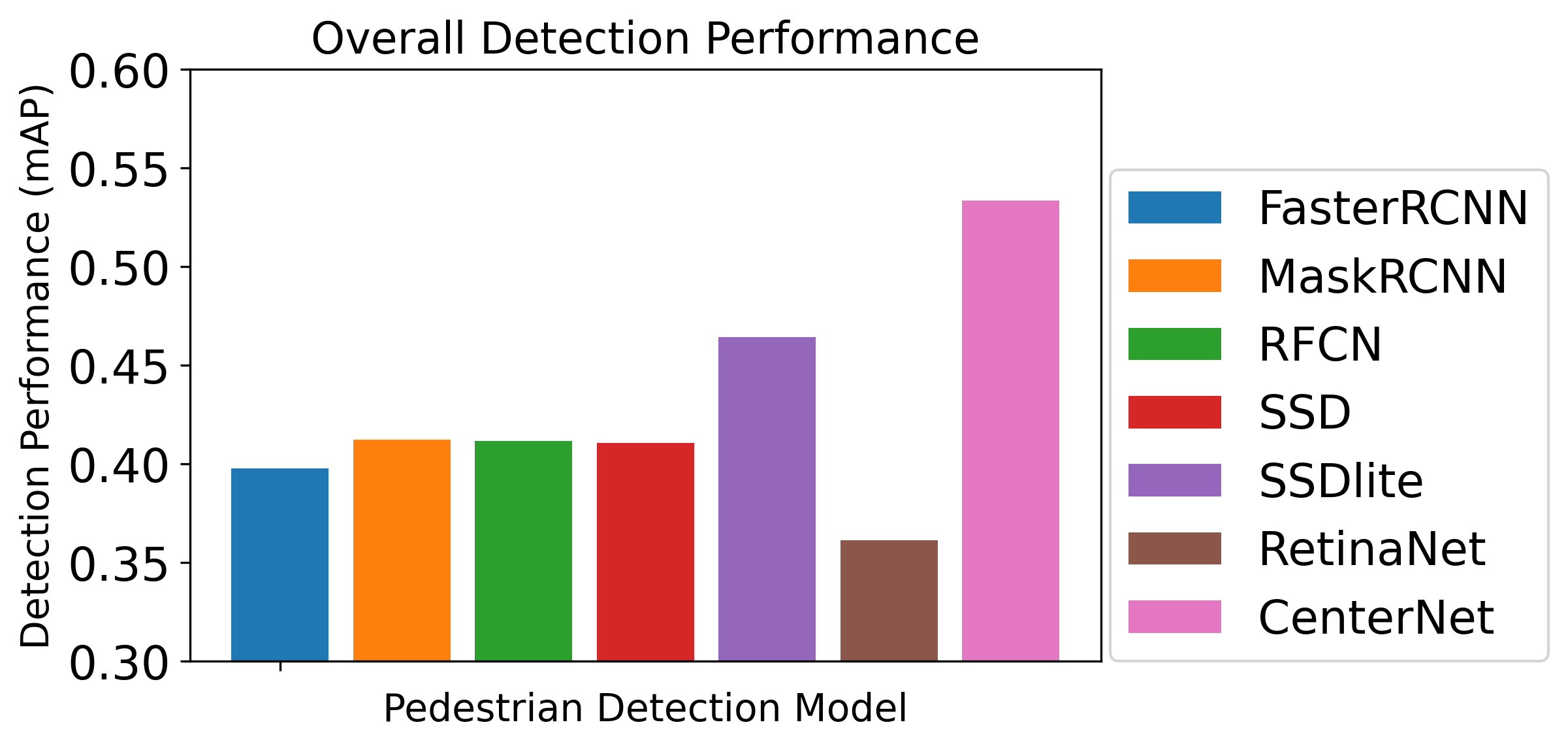}
\end{center}
  \caption{Overall Detection Performance. CenterNet has the greatest overall performance with a mAP of 0.533, followed by SSDlite (mAP = 0.464). RetinaNet has the lowest overall performance on the test dataset with a mAP of 0.361.}
\label{fig:overall mAP}
\end{figure}



\begin{figure*}
\begin{center}
\subfloat[]{\includegraphics[width=0.8\linewidth]{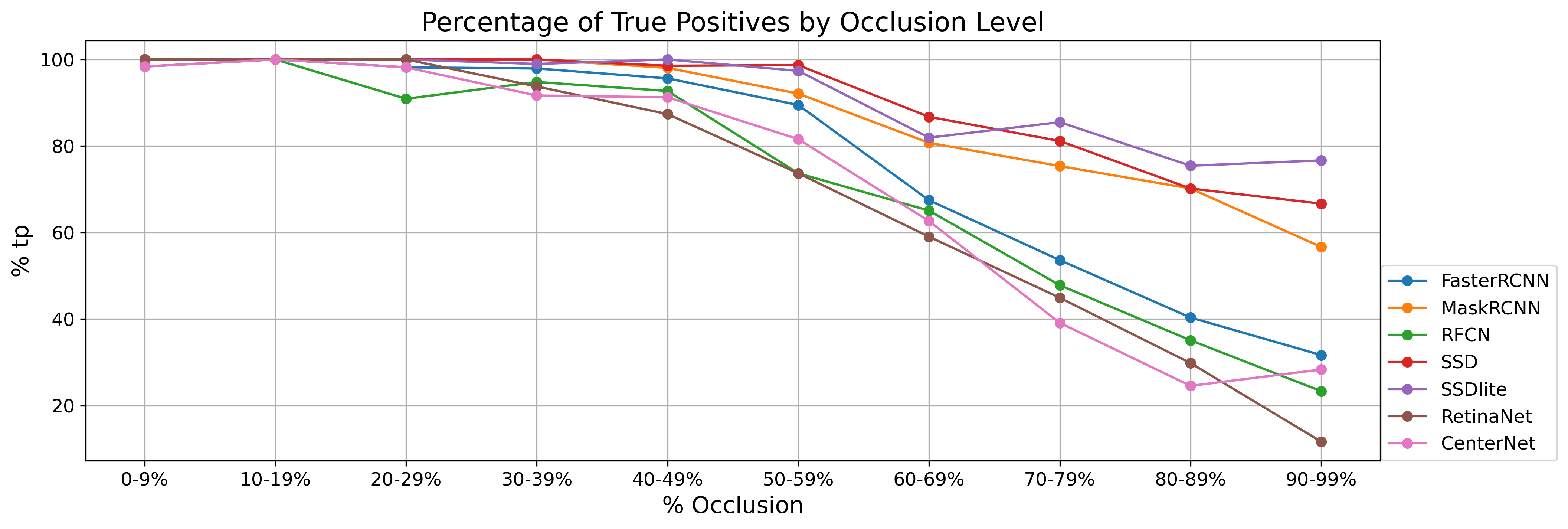}

\label{FigTP}}
\hfil
\subfloat[]{\includegraphics[width=0.8\linewidth]{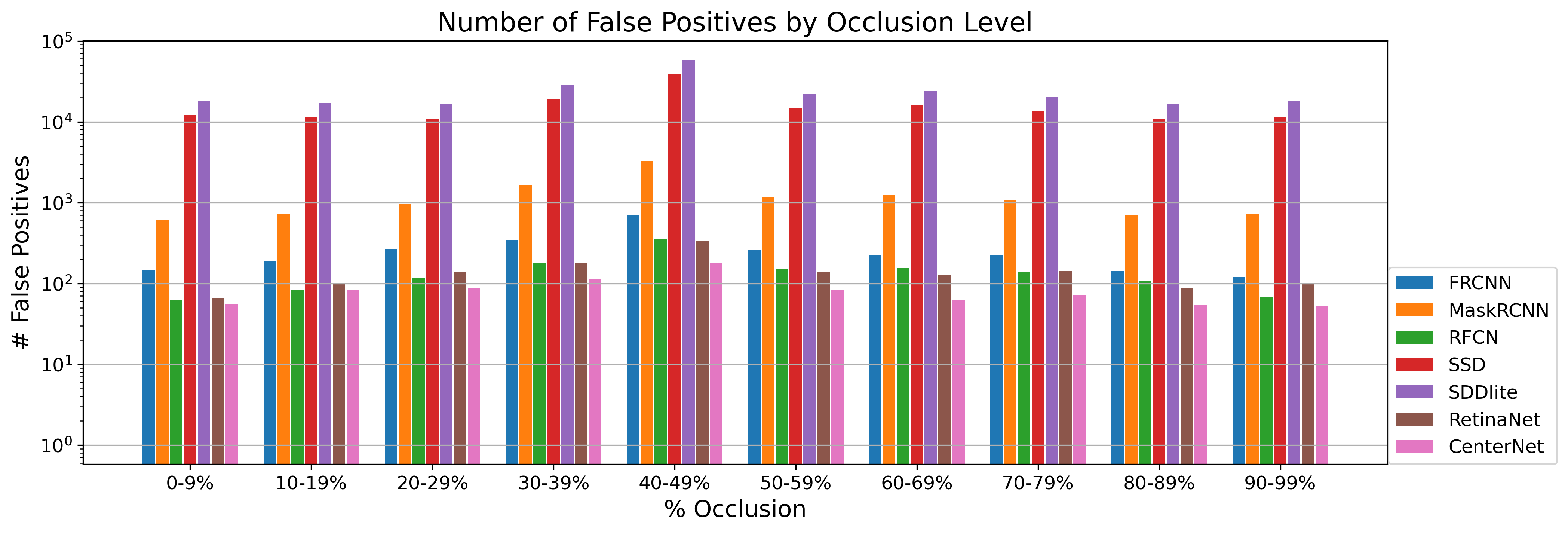}

\label{FigFPlog}}

\hfil

\subfloat[]{\includegraphics[width=0.8\linewidth]{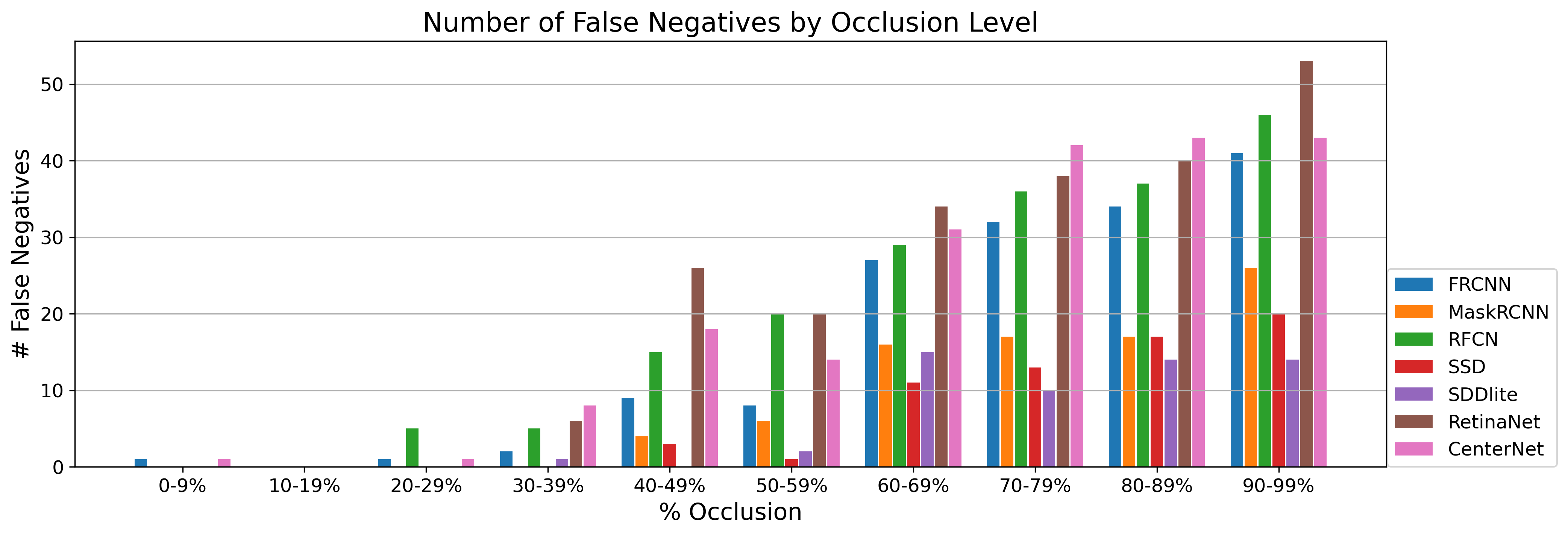}

\label{FigFN}}
\end{center}
\caption[True Positives, False Positives and False Negatives]{True Positives, False Positives and False Negatives. (a) displays the percentage of true positive detections by occlusion level for seven popular pedestrian detection models. (b) displays the number of false positives per occlusion level for each model. Note the logarithmic scale on the Y-axis. (c) displays the number of false negatives by occlusion level. } 
\label{fig:TP_FP_FN}
\end{figure*}

\begin{figure}[]
\begin{center}
   \includegraphics[width=\linewidth]{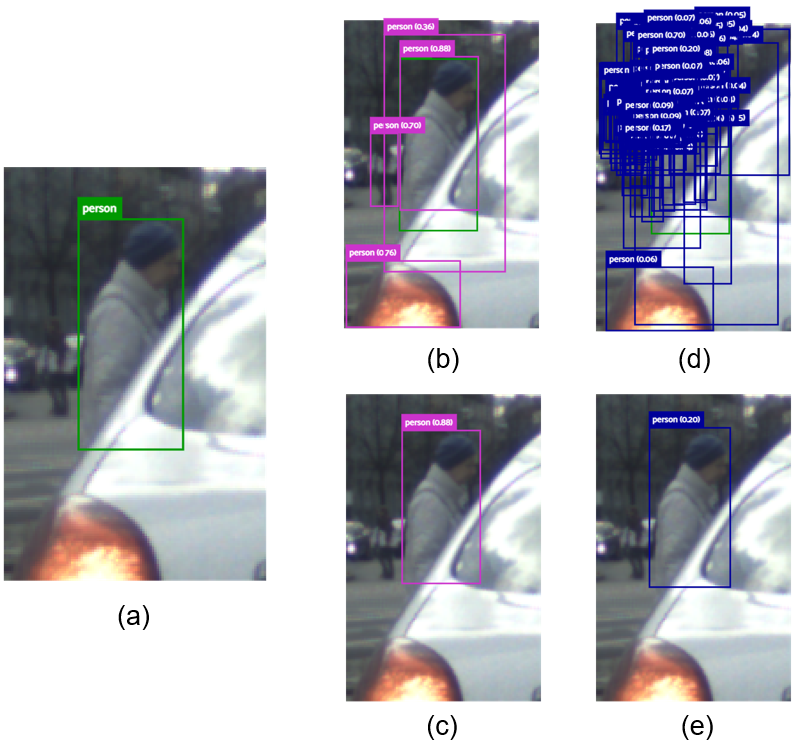}
\end{center}
   \caption{FasterRCNN vs. SSD. Detection performance is compared for a two stage network, FasterRCNN vs. a one stage network, SSD for an occluded pedestrian. The ground truth is shown in (a). FasterRCNN generates 4 proposals (b), 1 true positive detection with 88\% confidence (c), and 3 false positives. SSD generates 84 detections (d), 1 true positive with 22\% confidence (e), and 83 false positives.}
\label{fig:FRCNNvsSSD}
\end{figure}


\begin{figure}[t]
\subfloat[]{\includegraphics[width=\linewidth]{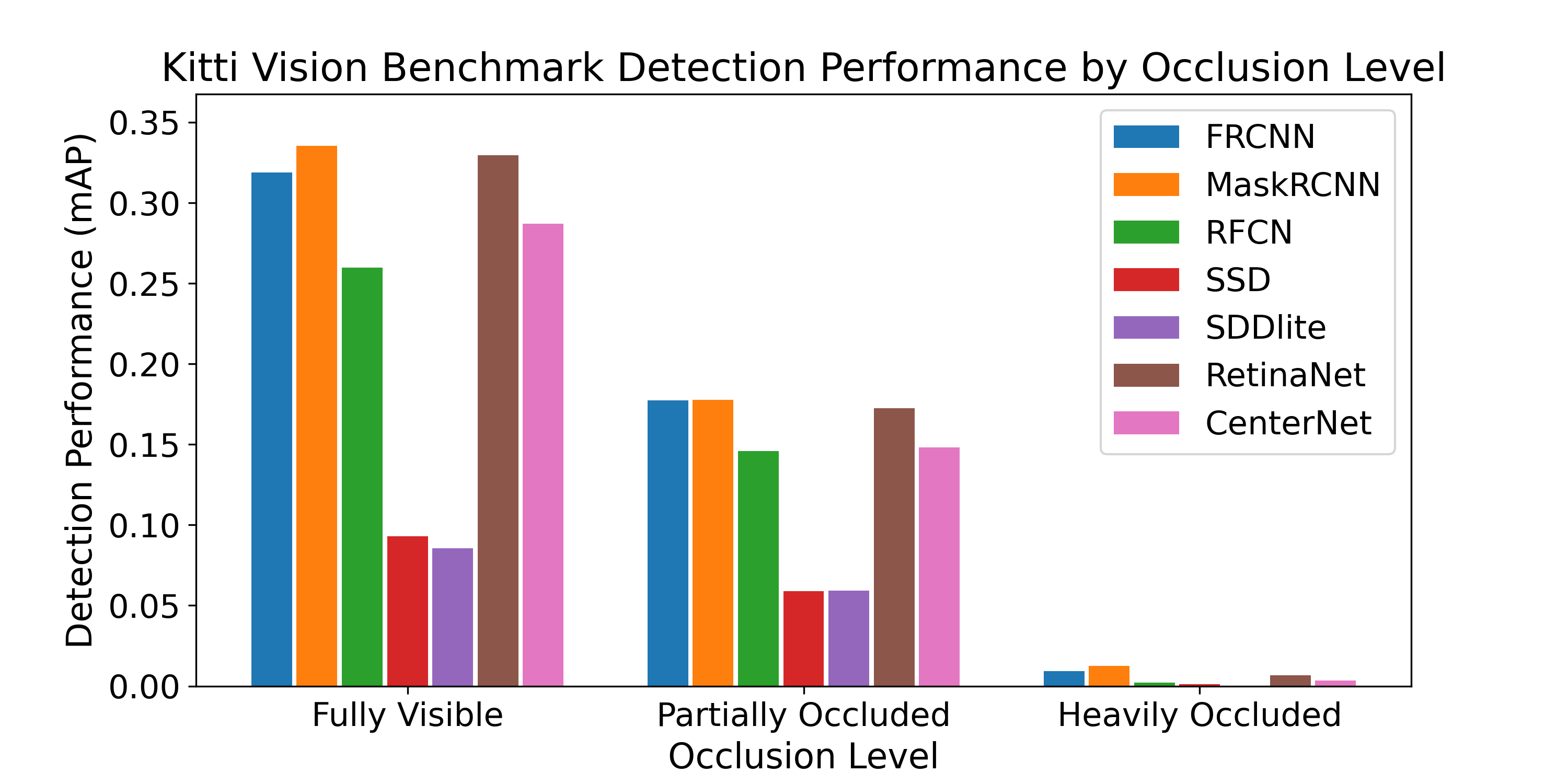}}

\subfloat[]{\includegraphics[width=\linewidth]{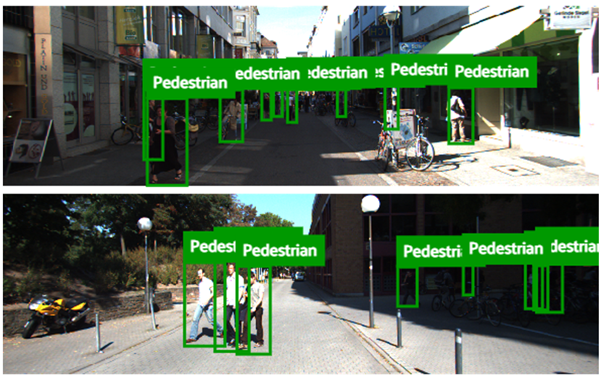}}

\caption{Detection performance by occlusion level using the Kitti Vision Benchmark, (a). Sample images from the Kitti Vision Benchmark are shown in (b). Note that pedestrian instances are impacted by a range of inhibiting factors in addition to partial occlusion such as object scale and lighting variations.}
\label{fig:kitti_occ_det}
\end{figure}


\begin{figure}[t]
\subfloat[]{\includegraphics[width=\linewidth]{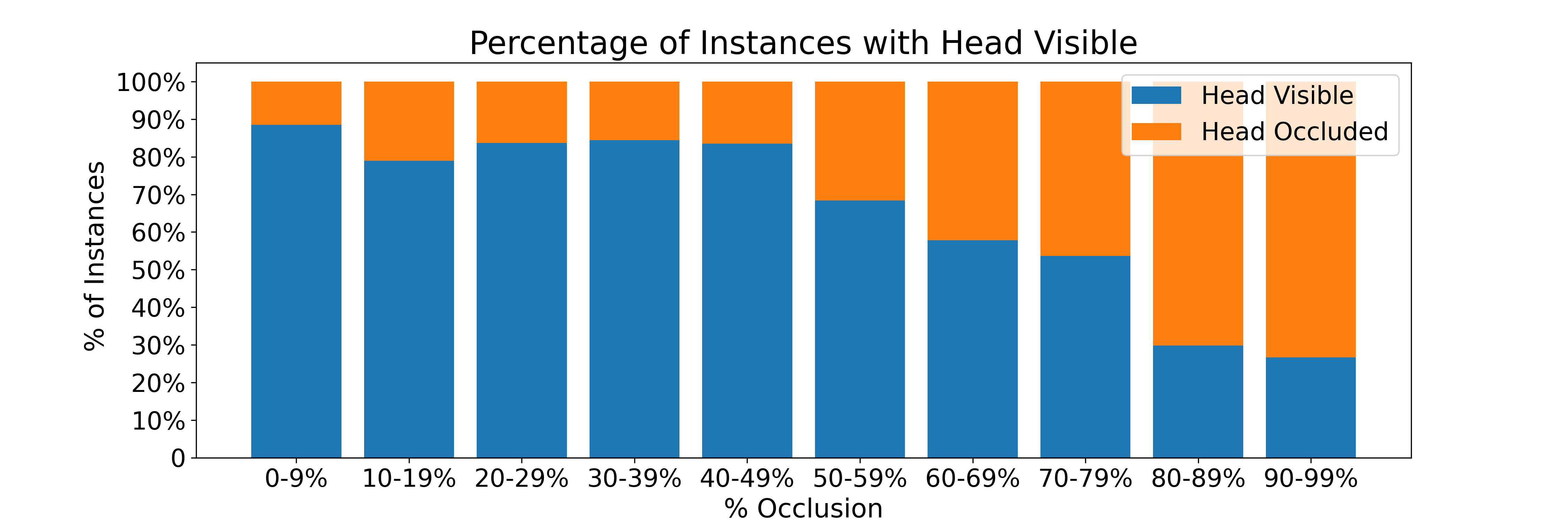}}

\subfloat[]{\includegraphics[width=\linewidth]{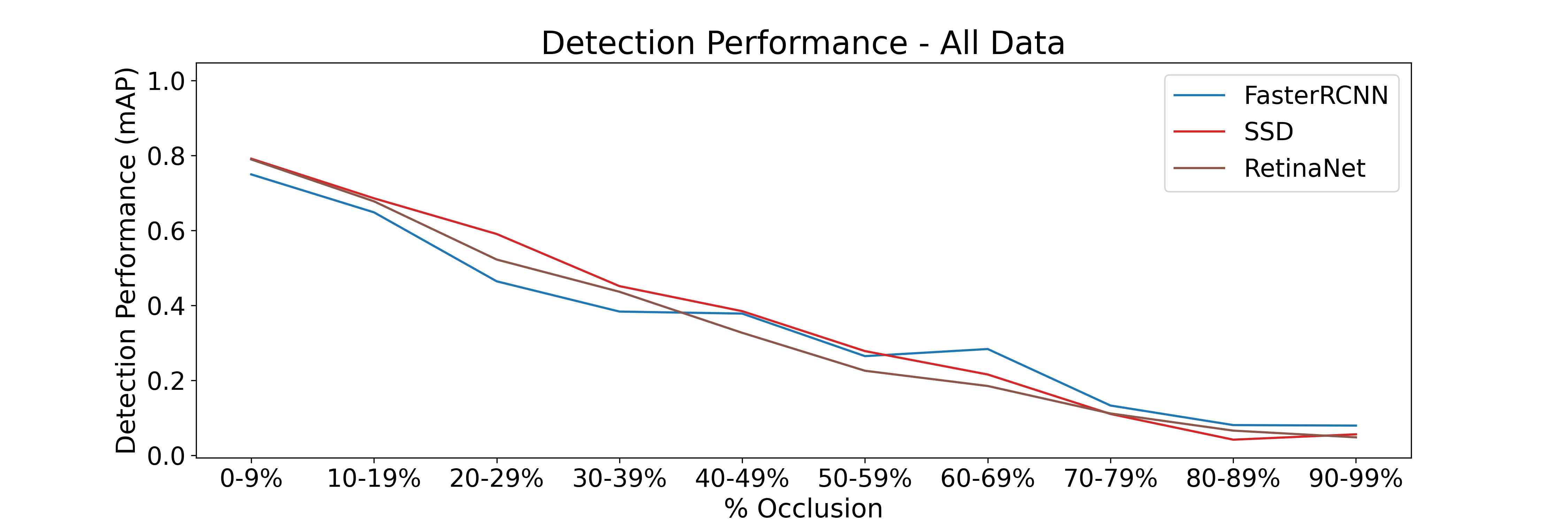}}

\subfloat[]{\includegraphics[width=\linewidth]{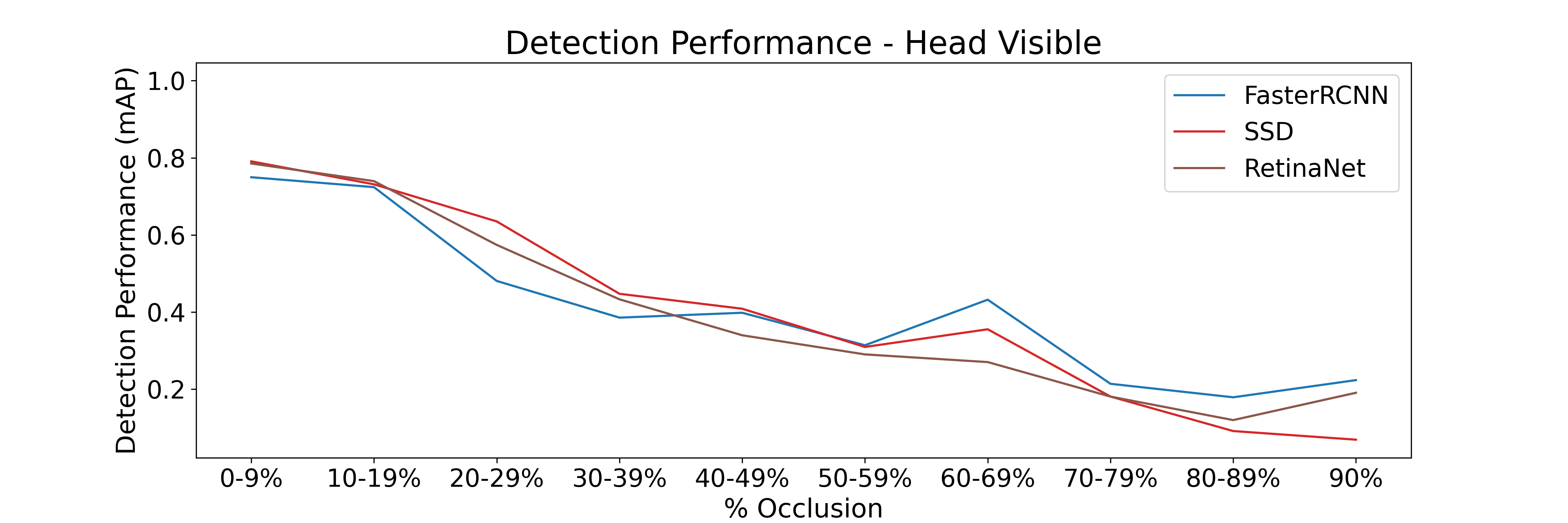}}

\subfloat[]{\includegraphics[width=\linewidth]{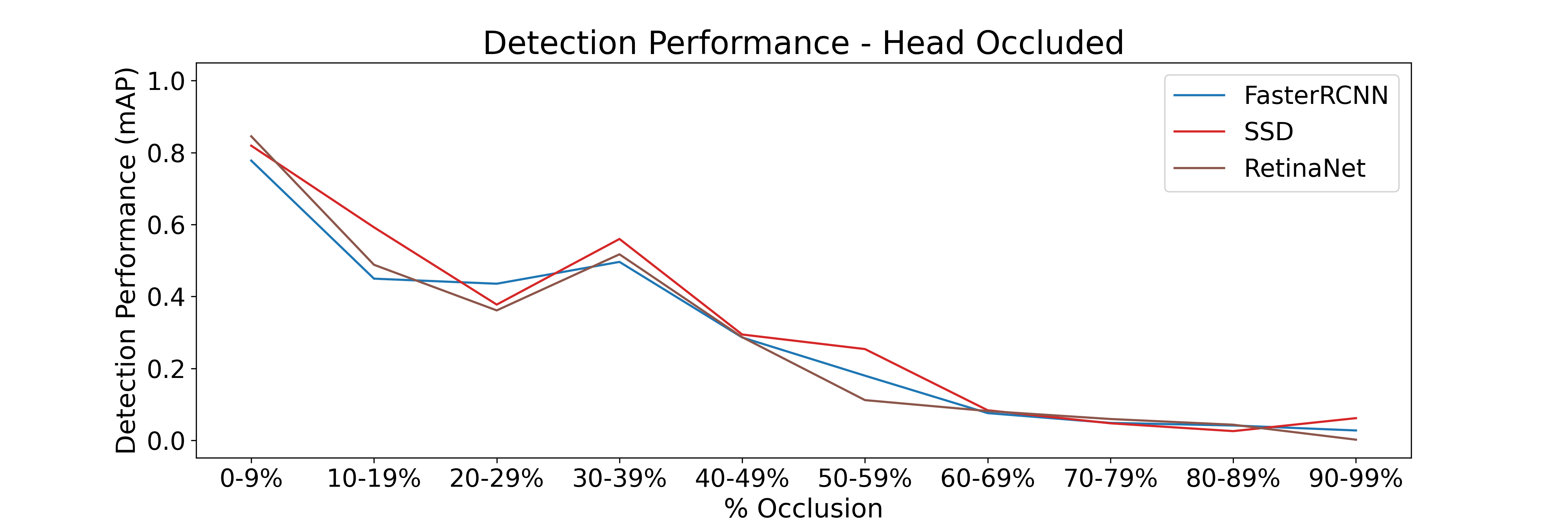}}

\caption{Analysis of data based on head visibility. (a) Dataset statistics based on head visibility. 820 total instances, 568 instances with head visible, 252 instances with head occluded. (b) Detection performance for FasterRCNN, SSD and RetinaNet for all data. (c) Detection performance for only images with head visible. Note, no occlusion level of more than 90\% possible with head visible. (d) Detection performance for only images with head occluded. Results demonstrate that, in general, detection performance degrades as occlusion level increases regardless of the presence of key semantic features such as a pedestrian's head.}
\label{fig:headvis}
\end{figure}

\section{Methodology}

A novel occluded pedestrian test dataset, containing 820 person instances in 724 images, has been created in order to characterize pedestrian detection performance across a range of occlusion levels from 0 to 99\% occluded. A diverse mix of images are used ensure that a wide variety of target pedestrians, pedestrian poses, backgrounds, and occluding objects are represented. The dataset contains both natural and superimposed occlusions in order to facilitate pedestrian detection characterization for the complete spectrum of occlusion levels from 0-99\%.

The dataset is sourced from three main categories of images: 
1) The "occluded body" subset of the partial re-identification dataset "Partial ReID" provided by Zheng \textit{et al} \cite{zheng2015partial},
2) The Partial ReID "whole body" subset \cite{zheng2015partial} with custom superimposed occlusions and
3) Images collated from publicly available sources including  \cite{braun2019eurocity}\cite{zhang2017citypersons}\cite{gilroy2021pedestrian}\cite{zhuo2018occluded}.
All images are annotated using the objective occlusion level classification method described in \cite{gilroy2022objective}. Occlusion level classification consists of the following steps: 1. Keypoint detection is applied to the input image in order to identify the presence and visibility of specific semantic parts for each pedestrian instance. 2. A visibility threshold is applied to identify occluded keypoints. 3. MaskRCNN is applied to define the pedestrian mask area and results are cross-referenced with detected keypoints to confirm which keypoints are occluded within the image. 4. Visible keypoints are then grouped into larger semantic parts and the total visible surface area is calculated using the 2D body surface area estimation method displayed in Figure \ref{fig:2DBSA} \cite{gilroy2021pedestrian}\cite{gilroy2022objective}.
Complex cases at very high occlusion rates were manually verified using the method of 2D body surface area estimation presented in Figure \ref{fig:2DBSA}. Each occlusion level contains a minimum of 55 pedestrian instances. A sample of the test dataset and dataset statistics by occlusion level can be seen in Figure \ref{fig:ds_sample} and Figure \ref{fig:ds_stats} respectively.

\subsection{Pedestrian Detection Models}

Performance characterization was carried out on seven popular pedestrian detection models. All models use publicly available pretrained weights from two popular model zoos \cite{voxel51_mz}\cite{torchvision_mz} and are trained using the COCO “train 2017” dataset \cite{lin2014microsoft}. An overview of the pedestrian detection models can be seen in Table \ref{table:models}.
The pedestrian detection models chosen for characterization can be divided into 3 categories: Two-Stage Frameworks, One-Stage Frameworks and Keypoint Estimation.
Two-stage frameworks such as FasterRCNN\cite{ren2016faster}, MaskRCNN\cite{he2017mask} and R-FCN\cite{dai2016r} apply two separate networks to perform classification. One network is used to propose regions of interest and a dedicated second network performs object detection \cite{chen2021deep}. One-stage frameworks such as RetinaNet\cite{lin2017focal}, SSD\cite{liu2016ssd} and SSDLite\cite{howard2019searching}\cite{sandler2018mobilenetv2} attempt to reduce computation and increase speed by performing object detection using a single feed forward convolutional network that does not interact with a region proposal module. RetinaNet also implements a novel method of “focal loss” which is used to reduce the imbalance between foreground and background classes during training with a view to increasing detection precision.
CenterNet\cite{zhou2019objects} takes an alternative approach based on keypoint estimation. Objects are represented as a single point at their bounding box center identified by a heat map generated using a fully convolutional network. Other object features such as object size, orientation and pose are then regressed directly from the image features at the center location. CenterNet has been shown to outperform a number of state of the art one-stage and two-stage algorithms in terms of a speed-accuracy trade off by maintaining an efficient network architecture\cite{zhou2019objects}.

\subsection{Experiments}

Detection performance is analyzed for the complete test dataset, and for each occlusion range from 0-9\% to 90-99\%, for pedestrian detection models to assess the impact of progressive levels of occlusion on the detectability of pedestrians. Analysis is carried out using Voxel51\cite{moore2020fiftyone} and the COCO style evaluation metric Mean Average Precision (mAP). Mean Average Precision is a popular and rigorous metric for object detection that calculates the Average Precision (AP) for a range of Intersection over Union (IoU) values from 0.5 to 0.95 with a step size of 0.5 and produces the mean value\cite{lin2014microsoft}. 
A summary of the results are shown in Figure \ref{fig:OccVsDet}, Figure \ref{fig:overall mAP} and Figure \ref{fig:TP_FP_FN}.
All models are also characterized using the Kitti Vision Benchmark \cite{geiger2012we} in order to compare and demonstrate the advanced analysis capabilities provided by the proposed benchmark.
Results on the Kitti Vision Benchmark are shown in Figure \ref{fig:kitti_occ_det}.


\section{Results and Analysis}

Results demonstrate that pedestrian detection performance (mAP) declines as the level of pedestrian occlusion increases, Figure \ref{fig:OccVsDet}. The number of false negative detections increase as occlusion level increases, Figure \ref{fig:TP_FP_FN}(c) and in general, the number of true positive detections begin to significantly decrease as occlusion level increases for pedestrians more than 50\% occluded, Figure \ref{fig:TP_FP_FN}(a). 
As shown in Figure \ref{fig:overall mAP}, of the seven popular pedestrian detection models analyzed, CenterNet \cite{zhou2019objects} has the greatest overall detection performance for partially occluded pedestrians with an overall mAP of 0.533, followed by SSDLite \cite{howard2019searching}\cite{sandler2018mobilenetv2} with a total dataset mAP of 0.464. The strategy employed by CenterNet of first identifying the bounding box centre using a keypoint heatmap and then predicting object size and bounding box dimensions relative to the centre point has demonstrated the highest precision bounding boxes for both fully visible pedestrians and for pedestrians up to 80\% occluded, Figure \ref{fig:OccVsDet}. MaskRCNN \cite{he2017mask} has the greatest detection performance for pedestrians occluded more than 80\%, Figure \ref{fig:OccVsDet}. RetinaNet \cite{lin2017focal} is the lowest performing overall on the test data with a mAP of 0.361. RetinaNet’s true positive detections begin to significantly degrade when pedestrians are more than 30\% occluded and this model has the highest number of false negatives for pedestrians more than 30\% occluded, Figure \ref{fig:TP_FP_FN}(a) and \ref{fig:TP_FP_FN}(c). 
Single Shot Detectors, SSD \cite{liu2016ssd} and SSDLite \cite{howard2019searching}\cite{sandler2018mobilenetv2} have the highest number of true positive detections at high levels of occlusion, Figure \ref{fig:TP_FP_FN}(a), and maintain a very high level of true positive detections up to 60\% occlusion, however their false positive rate is in the region of 100 times larger than popular two stage detectors such as FasterRCNN \cite{ren2016faster} and RFCN \cite{dai2016r} and approximately 16 times larger than MaskRCNN \cite{he2017mask}, Figure \ref{fig:TP_FP_FN}(b). Unlike false negatives, the number of false positives per image does not appear to be significantly impacted by the occlusion level as these are not typically related to the target pedestrian in an image. SSDlite \cite{howard2019searching}\cite{sandler2018mobilenetv2} outperforms SSD \cite{liu2016ssd} for almost all levels of occlusion despite having a higher number of false positive detections. MaskRCNN \cite{he2017mask} has a higher percentage of true positives than Faster RCNN \cite{ren2016faster} for pedestrians over 40\% occluded, however, it has around 4 times more false positive detections for the same data, Figure \ref{fig:TP_FP_FN}. Mask RCNN, RFCN and SSD all have similar overall performance on the test dataset, however, MaskRCNN and RFCN have a higher detection performance than SSD for pedestrians that are more than 60\% occluded, Figure \ref{fig:OccVsDet}.

Figure \ref{fig:FRCNNvsSSD} compares the output from a two stage detector, FasterRCNN, with a one stage detector, SSD, for an occluded pedestrian. Two stage detectors first generate key regions of interest before applying object detection, one stage detectors directly apply object detection to the entire image. Figure \ref{fig:FRCNNvsSSD} demonstrates that for the same image, FasterRCNN produces 4 detection outputs (1 true positive with 88\% confidence and 3 false positives), Figures \ref{fig:FRCNNvsSSD}(b) and \ref{fig:FRCNNvsSSD}(c), whereas SSD produces 84 detection outputs (1 true positive with 20\% confidence and 83 false positives), Figures \ref{fig:FRCNNvsSSD}(d) and \ref{fig:FRCNNvsSSD}(e). This indicates that all detection models must not be treated equally in the design of a pedestrian detection system. The characteristics and weaknesses of each detection model identified through robust performance characterization, must be taken into account further downstream in the object detection system, as some model outputs may be less reliable than others for safety critical systems.

\subsection{Benchmark Comparison}

Although a number of datasets contain occlusion labels to indicate the level of occlusion, current benchmarks are not designed for thorough characterization of partially occluded pedestrian detection performance. Each benchmark varies greatly in their definition of the occurrence and severity of occlusion and each benchmark uses different but highly subjective methods of occlusion level annotation, Table \ref{tab:datasets} \cite{gilroy2022objective}. In addition, many pedestrian instances are impacted by multiple additional inhibiting factors, making it difficult to determine if the contributing factor to non-detection is occlusion level alone. Algorithm performance can still be compared using the current state of the art, however users are unable to determine with any certainty if any non-detection is the result of occlusion or one of many other inhibiting factors such as object scale, distance from camera, adverse weather and lighting variations. This also makes it very difficult to accurately compare algorithm performance across multiple benchmarks. 

If we take the popular KITTI Vision Benchmark as an example. Images are annotated for three levels of occlusion: "Fully Visible", "Partially Occluded", "Difficult to See". Images are captured using a wide angle lens and contain many contributing factors to non-detection in addition to occlusion as shown in Figure \ref{fig:kitti_occ_det}(b).
The dataset is split into three test subsets in order to characterize pedestrian detection models by occlusion label:
1.) Images that only contain pedestrians tagged as "Fully Visible" (\textit{1669 Instances in 1242 Images}); 2.) Images that only contain pedestrians tagged as "Partially Occluded" (\textit{236 Instances in 216 Images}) and 3.) Images that only contain pedestrians tagged as "Difficult to See" (\textit{208 Instances in 158 Images}). Sitting persons and persons on bicycles are included for test purposes in cases where they have a suitable occlusion label.
Pedestrian detection performance is then assessed on each of the three subsets as shown in Figure \ref{fig:kitti_occ_det}(a). Results demonstrate that performance degrades for each broad, more complex data subset and MaskRCNN \cite{he2017mask} has the greatest overall performance on the Kitti Vision Benchmark data. However, partial occlusion can not be concluded as the only contributing factor to non-detection as many pedestrian instances have a number of additional inhibiting factors such as object scale, distance from camera and lighting variations. 

In contrast to this, the proposed benchmark facilitates detailed, objective and repeatable characterization of pedestrian detection performance specifically for partially occluded pedestrians across the complete range of occlusion levels from 0-99\%, Figure \ref{fig:OccVsDet}.

\subsection{Key Semantic Parts}




Further analysis has been carried out to determine the impact that the visibility of a pedestrian’s head has on the detection of occluded pedestrians. The dataset was split into two subsets: 1) Only images where the target pedestrian's head is visible and 2) Only images where the target pedestrian's head is occluded. Of the 820 pedestrian instances, the target pedestrian's head is visible in 582 instances and is occluded in 252 instances. Figure \ref{fig:headvis}(a) displays the percentage of pedestrian instances with their head visible across each of the occlusion levels. Three pedestrian detection models, FasterRCNN, RetinaNet and SSD were then tested on both data subsets across the occlusion range. Experiments demonstrate that, regardless of whether a pedestrians head is visible, a distinct declining profile in detection performance is observed as pedestrian occlusion level increases, Figures \ref{fig:headvis}(b), \ref{fig:headvis}(c) and \ref{fig:headvis}(d). Results indicate that the detection models under test are not biased towards head visibility for the classification of partially occluded pedestrians.

\section{Conclusion}
Detection of partially occluded pedestrians remains a persistent challenge for driver assistance systems and autonomous vehicles. Current methods of characterizing detection performance for partially occluded pedestrians have been broad, subjective, and inconsistent in their definition of the level of occlusion.
This research presents a novel test benchmark for the detailed, objective analysis of pedestrian detection models for partially occluded pedestrians. Detection performance is characterized for seven popular pedestrian detection models across a range of occlusion levels from 0-99\%. The proposed benchmark focuses specifically on the complex issue of partial occlusion and facilitates more objective, repeatable and fine grained analysis than the current state of the art.
Results demonstrate that pedestrian detection performance experiences a negative correlation to increases in occlusion level as the visibility of a pedestrian is incrementally reduced. An increase in the number of false negative detections is observed as occlusion level increases and the percentage of true positive detections significantly degrade for pedestrians who are more than 50\% occluded.
Further analysis demonstrates that not all pedestrian detection models should be treated equally within an object detection system. The speed vs. accuracy trade-off, encouraged by the near real-time requirements of autonomous vehicles, can result in high levels of false positive detections and lower detection confidence at progressive levels of pedestrian occlusion, particularly when using single stage detection models.
Thorough objective characterization of pedestrian detection models at the design stage will improve the performance of object detection systems by calibrating the priority of detections in scenarios where known weaknesses can occur. System improvements may be gained through the use of an occlusion-aware step in the object detection pipeline to inform the priority of camera-based detections in sensor fusion networks for SAE level 4 and level 5 autonomous vehicles. In this manner, any reduction in performance at high occlusion levels can be mitigated in the design of the overall system to increase the safety of vulnerable road users and improve the efficiency of path planning based on environment detection.
Widespread use of the proposed benchmark will result in more objective, consistent and detailed analysis of pedestrian detection models for partially occluded pedestrians.


\ifCLASSOPTIONcaptionsoff
  \newpage
\fi


{\small
\bibliographystyle{IEEEtran}
\bibliography{refs}
}

\end{document}